\documentclass[12pt]{article}

\usepackage{amssymb}
\usepackage{amsmath}
\usepackage{amscd}
\usepackage{latexsym}
\usepackage{graphicx}

\usepackage{cite}
\usepackage{enumitem}

\newcommand{\be}{\begin{equation}}
\newcommand{\ee}{\end{equation}}
\newcommand{\Dlt}{\Delta}

\newcommand{\cA}{{\cal A}}

\newcommand{\ra}{\rightarrow}

\newcommand{\bt}{\beta}
\newcommand{\vp}{\varphi}

\newcommand{\lbd}{\lambda}

\begin{document}

\begin{center}

{\Large{\bf Quantification of emotions in decision making} \\ [5mm]

V.I. Yukalov } \\ [3mm]

{\it Bogolubov Laboratory of Theoretical Physics, \\
Joint Institute for Nuclear Research, Dubna 141980, Russia \\
and \\
Instituto de Fisica de S\~ao Carlos, Universidade de S\~ao Paulo, \\
CP 369,  S\~ao Carlos 13560-970, S\~ao Paulo, Brazil } \\ [2mm]
                          
{\bf e-mail}: yukalov@theor.jinr.ru

\end{center}

\vskip 2cm

\begin{abstract}

The problem of quantification of emotions in the choice between alternatives is 
considered. The alternatives are evaluated in a dual manner. From one side, they 
are characterized by rational features defining the utility of each alternative. 
From the other side, the choice is affected by emotions labeling the alternatives 
as attractive or repulsive, pleasant or unpleasant. A decision maker needs to make 
a choice taking into account both these features, the utility of alternatives and 
their attractiveness. The notion of utility is based on rational grounds, while 
the notion of attractiveness is vague and rather is based on irrational feelings. 
A general method, allowing for the quantification of the choice combining rational 
and emotional features is described. Despite that emotions seem to avoid precise 
quantification, their quantitative evaluation is possible at the aggregate level. 
The analysis of a series of empirical data demonstrates the efficiency of the 
approach, including the realistic behavioral problems that cannot be treated by 
the standard expected utility theory. 
 
\end{abstract}

\vskip 1cm

{\parindent=0pt
{\bf Keywords}: emotions in decision making, quantification of emotions, 
behavioral probability, dual choice, affective computing, artificial intelligence 
}

\vskip 3mm

{\parindent=0pt
{\bf Declarations} 
Funding: Not applicable,
Conflicts of interest: Not applicable,
Availability of data and material: Not applicable,
Code availability: Not applicable 
}

\vskip 2mm

ORCID: 0000-0003-4833-0175

\newpage

\section{Introduction}

The problem of making a choice between alternatives is a core of decision theory and 
its numerous applications in economics, finances, and the operation of intelligence, 
whether artificial or human. The most well developed procedure of decision making is 
based on the expected utility theory formalized by von Neumann and Morgenstern (1953). 
However, as is well known, it is rather a rare occasion when decisions are made on the 
basis of purely rational grounds estimating the alternative utility. Almost always the
choice is essentially affected by emotions, and humans do not strictly follow the
prescriptions of the expected utility theory, which results in numerous paradoxes and 
often does not allow even for qualitative predictions. To take into account behavioral 
effects related to the influence of emotions and other subjective biases, various 
so-called non-expected utility theories were suggested by replacing the expected utility 
with specially constructed functionals invented for the purpose of a posteriori 
interpretation of one or just a few phenomena. A list of such non-expected utility 
theories can be found in the review by Machina (2008).     

However, non-expected utility theories are descriptive requiring fitting of several 
parameters from particular experimental data. In addition, spoiling the structure of 
the expected utility leads to the appearance of inconsistences and new paradoxes 
producing more problems than it resolves (Safra and Segal 2008; Birnbaum 2008; 
Al Najjar and Weinstein 2009, 2009). 

The major problem in describing real-life decision making is caused by the 
difficulty of quantifying such behavioral phenomena as emotions. This is because
subjective emotions are not precisely defined in explicit mathematical terms, 
contrary to such a crisp notion as utility that can be evaluated on rational grounds. 
Therefore emotions increase the uncertainty that always exists in any choice, when 
decision makers evaluate the features of the given alternatives (Scherer and Moors 2019). 
When analyzing alternatives, decision makers experience different feelings, emotions, 
and subconscious intuitive movements (Kahneman 1982; Picard 1997; Minsky 2006; 
Plessner et al 2008). This is why, even choosing between seemingly well formulated 
lotteries, humans often do not obey the normative prescriptions of utility theory, but 
make decisions qualitatively contradicting the latter (Kahneman and Tversky 1979).    

It is important to differentiate two sides in the problem of emotion quantification.
One side is the assessment of emotions experienced by a subject as reactions on external
events, e.g. hearing voice or looking at pictures. The arising emotions can include
happiness, anger, pleasure, disgust, fear, sadness, astonishment, pain, and so on. The 
severity or intensity of such emotions can be estimated by studying the expressive forms 
manifesting themselves in motor reactions, such as facial expressions, pantomime, and 
general motor activity, and by measuring physiological reactions, such as the activity 
of the sympathetic and parasympathetic parts of the autonomic nervous system, as well 
as the activity of the endocrine glands. Vegetative manifestations of emotions can be
noticed by studying changes in the electrical resistance of the skin, the frequency and 
strength of heart contractions, blood pressure, skin temperature, hormonal and chemical 
composition of the blood, and like that. Several  methods of appraising particular 
emotions in separate setups have been considered (Amjadzadeh and Ansari-Asl 2017;
Vartanov and Vartanova 2018; Scherer and Moors 2019; Vartanov et al. 2020, Wang et al. 2021).  

The other, principally different, side of emotion evaluation concerns the study of the
influence of emotions on taking decisions by subjects. It is generally accepted that 
human decisions are not purely rational, but emotions do play a great role in decision 
making. However the quantitative influence of emotions on the process of decision making  
remains yet an unsolved problem.  

The present paper studies the second problem: How it would be possible to assess the 
influence of emotions on decisions taken by humans? We do not consider the somatic or
physiological effects produced by emotions, but we aim at analyzing how subjective 
emotions, arising in the process of decision making, influence the resulting decisions. 

Subjectivity in decision making arises because of uncertainty in the suggested choice. 
This uncertainty can be of dual nature. From one side, there is the usual probabilistic 
uncertainty based on deliberations related to the alternative utility. From the other 
side, there is an uncertainty in the choice due to the subjective feelings that are not 
regulated by rational rules. Emotions can be separated into three classes. One class 
contains, loosely speaking, positive, features, such as "good", "pleasant", "attractive" 
and like that. The second class is composed of negative characterizations, such as "bad", 
"unpleasant", "repulsive" etc. And the third class is intermediate, comprising neutral 
definitions expressing indifference with respect to the alternatives under consideration. 
 
Despite the subjectiveness of emotions, their influence in the choice between 
alternatives sometimes can be quantified. Of course, this looks to be impossible 
for a particular decision maker and for each separate choice procedure. Yet, it turns 
out that quantification is admissible at the aggregate level for a typical decision maker 
representing the average characteristics of a large group of decision makers.   
         
The formulation of explicit mathematical rules allowing for the selection of an optimal 
alternative under vague uncertainty due to the influence of emotions, is not merely 
useful for characterizing human decision making, but it is compulsory for the 
realization of affective computing (Picard 1997) and for overcoming the challenge 
of creating artificial intelligence (Russel and Norvig 2016; Poole and Mackworth 2017; 
Neapolitan and Jiang 2018). The achievement of human-level machine intelligence is a 
principal goal of artificial intelligence since its inception. 

The process of decision making, actually, consists of two sides that can conditionally 
be named rational and irrational. The rational side describes the comparative usefulness 
of the considered alternatives, while the irrational side is due to emotions making the 
decision process less predictable. It is the rational-irrational duality that makes 
the quantification of the decision-making process so difficult.   

The distinction between rational and irrational has been extensively discussed in 
the literature on dual processes (Sun 2002; Paivio 2007; Evans 2007; Stanovich 2011; 
Kahneman 2011) according to which the procedure of taking decisions in human brains 
can be treated as a result of two different processes that can be called rational 
(logical, controlled, regulated, deterministic, slow, defined by clear rules) and 
irrational (intuitive, uncontrolled, emotional, random, fast, defined in a fuzzy 
manner). These processes may proceed in parallel or in turn, but in any case they 
act differently (Milner and Goodale 2008; Kahneman 2011). 

It is important to stress that the differentiation of mental processes onto rational 
and irrational has the meaning for the moment of taking a decision (Ariely 2008). 
It is a psychological distinction but not a philosophical one. It is clear that 
giving a specially invented philosophical definition it is straightforward to include 
afterwards all intuitive and emotional effects into the rank of rational just giving 
a definition that rational is all what leads to the desired goal. Then illogical 
uncontrolled feelings that occasionally lead to the goal should be termed rational, 
and vice versa logical conclusions that occasionally miss the goal should be named 
irrational (Searle 2001; Julmi 2019). The philosophical definition of rational has 
the meaning only afterwards, when the goal has been reached. Only then it becomes 
clear what was leading to the goal and what was not. 

Moreover, the philosophical definition of rational as what leads to the goal is 
ambiguous. For instance, assume that your goal is to become rich. The easiest way 
to become rich is to steal. Hence to steal is rational. But then you are caught by 
police and put into jail. To be jailed was not your goal. Hence to steal is not 
rational. So it is not clear, is it rational or not, while from the psychological 
point of view there is no ambiguity. An action that is logically and explicitly 
formulated is psychologically rational. The psychological definition is based on 
real physiological processes in the brain, while the philosophical definition is 
not uniquely defined and depends on interpretations.

In what follows, we distinguish rational from irrational as it is accepted in 
decision making, where rational is what can be explicitly formulated, based on 
clear rules, deterministic, logical, prescriptive, normative, while irrational 
is the opposite to rational (Ariely 2008; Zafirovski 2012), being intuitive, 
uncontrolled, emotional, random, defined in a fuzzy way.  

The dual nature of decision making, comprising the rational-irrational duality, 
suggests that this duality could be mathematically represented by a theory that 
naturally includes some kind of duality in its basis. The proper candidate
for this could be quantum theory, with its particle-wave duality. A consistent
approach realizing this analogy, by treating decision making as the procedure of
quantum measurements, is the recently developed Quantum Decision Theory
(Yukalov and Sornette 2008, 2009, 2011, 2014, 2016, 2018; Yukalov 2020, 2021). 
 
However, mathematical techniques of quantum theory are not customary for the majority
of people. Therefore it would be desirable to develop a theory that could incorporate 
the achievements of quantum decision theory at the same time avoiding mathematical 
complications of quantum techniques and the language of quantum theory so unfamiliar 
for the majority of researchers. The development of such an approach and its farther 
elaboration is the goal of the present paper. Specifically, the new results of the 
present article are as follows. 

(i) The axiomatic formulation of dual decision theory, taking account of 
cognition-emotion duality, that is rational-irrational duality in decision making, 
and comprising the main points of quantum decision theory, but without involving 
any quantum formulae. 

(ii) The derivation, without appealing to quantum theory, of the non-informative 
prior estimate for attraction factor measuring the typical influence of emotions on 
the process of decision making. 

(iii) Illustration of a simple rule for distinguishing emotionally attractive and 
repulsive characteristics of the considered alternatives in the case of highly 
uncertain lotteries of the Kahneman-Tversky type. 

(iv) Analysis of empirical data confirming that the typical influence of emotions 
in decision making composes $25 \%$.

The plan of the paper is as follows. The approach to be formulated possesses two 
major features. First, it is probabilistic, which requires to define the corresponding 
probability measure. Second, it is dual, aiming at taking into consideration rational
as well as irrational characteristics of alternatives. In Sec. 2, the rules for 
defining the rational probabilistic choice, describing the utility of the alternatives, 
is formulated. In Sec. 3, it is shown how it is possible to characterize the emotional 
attractiveness of alternatives. In Sec. 4, the behavioral probability is defined, 
combining the rational utility measure and the irrational emotional characteristic, 
called attraction factor. The properties of the attraction factor are considered in 
Sec. 5, where the typical value of the attraction factor is found to be $1/4$, which 
is called the quarter law. This value allows for the estimation of non-informative 
priors for the attraction factors describing the influence of emotions at the aggregate 
level. Section 6 shows how the attraction factors for multiple alternatives can be 
estimated. Section 7, by the example of binary decision tasks, illustrates that the
definition of attraction factors, generally speaking, is contextual. This is because
the attraction factors can contain parameters whose values need to be adjusted for 
describing a particular set of decision problems, which limits their use for other
sets of decision tasks. In Sec. 8, the method of defining the attraction factor 
structure in the case of two alternatives with equal or very close utilities is
described. Section 9 considers difficult choice tasks in the case of the Kahneman-Tversky
lotteries, whose utilities are either exactly equal or very close to each other, 
so that the standard utility theory is not applicable. A method is suggested estimating
the quality of the lotteries and their attractiveness and giving good quantitative 
predictions at the aggregate level. In Sec. 10, the analysis of a large set of lotteries 
is given demonstrating the validity of the quarter law at the aggregate level. Section 11 
concludes.

\section{Probabilistic uncertainty}

The main task of decision theory is to describe the process of choice between a given 
set of alternatives
\be
\label{1}
 \cA = \{ A_n:\; n = 1,2,\ldots,N_A \} \;  .
\ee
Each alternative can be characterized from two sides, from the rational point of 
view of its usefulness and, from the other side, following irrational feelings and 
emotions. 

In this and the following sections, we describe a new approach to decision making,
taking into account the rational reasoning by estimating the utility of the considered 
alternatives as well as the presence of irrational emotions accompanying the choice.   

Even when there exist rational logical arguments explaining the utility of the 
given alternatives, not all subjects incline to prefer a single alternative, but 
always an alternative $A_n$, with a clearly defined utility, is selected only by 
a fraction $f(A_n)$ of decision makers (Slovic and Tversky 1974), which can be 
termed {\it rational fraction}. 

In the present section, the definition of the rational fraction is formulated and 
its properties are described. 

\vskip 2mm

{\bf Definition 1}. A rational fraction $f(A_n)$ is the fraction of decision makers 
that would choose the alternative $A_n$ provided their decisions would be based solely 
on rational grounds. The rational fraction is semi-positive and normalized,
\be
\label{2}
\sum_{n=1}^{N_A} f(A_n) = 1 \; , \qquad 0 \leq f(A_n) \leq 1 \;  .
\ee
The rational fraction represents the classical probability, with its standard 
properties, including the additivity with respect to mutually exclusive alternatives,
\be
\label{3}
 f \left( \bigcup_n A_n \right)  = \sum_n f(A_n) \; .
\ee
   
\vskip 2mm

{\bf Definition 2}. An alternative $A_1$ is called more useful than $A_2$ if and only 
if
\be
\label{4}
 f(A_1) > f(A_2) \;  .
\ee
Two alternatives, $A_1$ and $A_2$, are equally useful if and only if
\be
\label{5}
 f(A_1) = f(A_2) \;  .
\ee
The rational fraction $f(A_n)$ shows how useful the alternative is, because of which it 
can be called the {\it utility fraction}. 

\vskip 2mm

The ideas of decision theory ate typically illustrated by the choice between lotteries.
Let the alternatives be represented by the lotteries
\be
\label{6}
 A_n = \{x_i , \; p_n(x_i): ~ i = 1,2,\ldots , N_n \} \; , 
\ee
which are the probability distributions over payoffs $x_i$, with $p_n(x_i)$ being the 
payoff probabilities that can be either objective (von Neumannn and Morgenstern 1953)
or subjective (Savage 1954). The lottery utility is quantified by the expected utility
\be
\label{7}
U(A_n) = \sum_i u(x_i) p_n(x_i) \;  ,
\ee
where $u(x)$ is a utility function. More generally, it is possible to introduce a utility
functional 
$$
U(A_n) = \sum_i u(x_i) w(p_n(x_i)) \; ,
$$
with $w(p_n(x_i))$ being a postulated weighting function (Kahneman and Tversky 1979).

The rational fraction, associated with the expected utility, should satisfy the 
limiting conditions
$$
f(A_n) \ra 1 \; , \qquad U(A_n) \ra \infty \; ,
$$
\be
\label{8}
f(A_n) \ra 0 \; , \qquad U(A_n) \ra - \infty\;   ,
\ee
whose meaning is clear. An explicit form of the rational fraction can be done by 
the Luce rule (Luce 1959; Luce and Raiffa 1989) according to which, if an alternative 
$A_n$ is characterized by an attribute $a_n$, then the weight of this alternative can 
be defined as
\be
\label{9}
 f(A_n) = \frac{a_n}{\sum_{n=1}^{N_A}a_n} \qquad ( a_n \geq 0 ) \; .
\ee
When the expected utilities of all lotteries are semi-positive, the attribute values
can be defined by these utilities
\be
\label{10}
a_n = U(A_n) \; , \qquad U(A_n) \geq 0 \;   ,
\ee
while when the expected utilities are negative, the attribute values are defined by
the inverse quantities
\be
\label{11}
a_n = \frac{1}{|\; U(A_n)\;|} \; , \qquad U(A_n) < 0 \;  .
\ee
In the case of mixed utility signs, it is straightforward to shift the utilities by 
a minimal available wealth making these utilities semi-positive. 

\vskip 2mm

{\bf Definition 3}. When alternatives are represented by lotteries, the rational 
fractions can be defined as
\be
\label{12}
 f(A_n) = \frac{U(A_n)}{\sum_{n=1}^{N_A}U(A_n)} \qquad ( U(A_n) \geq 0 ) 
\ee
for semi-positive utilities and as
\be
\label{13}
f(A_n) = \frac{|\;U(A_n)\;|^{-1}}{\sum_{n=1}^{N_A}|\;U(A_n)\;|^{-1}} \qquad 
( U(A_n) < 0 ) 
\ee
for negative utilities. 

\vskip 2mm

Generally speaking, as the utility $U(A_n)$, one can imply either the standard
expected utility (von Neumannn and Morgenstern 1953), or a utility functional, 
for instance as is used in the prospect theory (Kahneman and Tversky 1979). It is
also possible to define the rational fraction as the minimizer of an information 
functional, as has been done for resolving the St. Petersburg paradox (Yukalov 2021).  

The so-defined rational fraction quantifies the fraction of decision makers choosing
an alternative being based only on rational arguments of the alternative utility. 
In other words, it is the probability that an alternative would be chosen by decision 
makers, provided they are purely rational.

\section{Emotional uncertainty}

In addition to the probabilistic uncertainty that can be quantified by the rational 
fraction, there exists an emotional uncertainty ascribing to the alternatives such 
vague emotional characteristics that do not seem to allow for a quantification. In the 
simplest case, these  characteristics can be separated into three classes of different 
quality. One quality class includes such specifications as "positive", "good", "pleasant", 
and "attractive", while the other is composed of such depictions as "negative", "bad", 
"unpleasant", and "repulsive". The third, intermediate class qualifies the related 
alternatives as "neutral" or "indifferent" with respect to their attractiveness. The 
principal question is how it would be possible to describe in mathematical terms and 
quantify these classes of emotional uncertainty? 

We shall denote the set of alternatives pertaining to the positive quality class 
as $\mathcal{A}_+$, while the set of alternatives pertaining to the negative quality 
class, as $\mathcal{A}_-$. The set of alternatives from the neutral quality class 
is denoted by $\mathcal{A}_0$. Let the emotional attractiveness of an alternative $A_n$ 
be represented by an {\it attraction factor} $q(A_n)$. For a positive, attractive 
alternative, the attraction factor is positive, for a negative, repulsive alternative, 
it is negative, and for a neutral alternative, it is zero. The absolute value of the 
attraction factor is limited by one.

\vskip 2mm

{\bf Definition 4}. The attraction factor pertaining to a positive, negative or neutral 
quality class, respectively, varies in the intervals
$$
0 < q(A_n) \leq 1 \qquad ( A_n \in \cA_+ ) \; ,
$$
$$
-1 \leq q(A_n) < 0 \qquad ( A_n \in \cA_- ) \; ,
$$
\be
\label{14}
q(A_n) = 0 \qquad ( A_n \in \cA_0 ) \;   .
\ee

\vskip 2mm

{\bf Definition 5}. An alternative $A_1$ is more attractive than $A_2$ if and only if 
\be
\label{15}
 q(A_1) > q(A_2) \;  .
\ee
Conversely, an alternative $A_2$ is more repulsive than $A_1$. Two alternatives are 
said to be equally attractive, or equally repulsive, if and only if
\be
\label{16}
 q(A_1) = q(A_2) \;  .
\ee

\vskip 2mm

Recall that quality, or attractiveness, is a vague subjective notion which can be 
interpreted as that the attraction factor is a random quantity varying in the frame of 
its quality class. The qualities "attractive" or "repulsive" are subjective, being 
associated with concrete decision makers. Moreover, they can change for the same 
decision maker taking decisions at different moments of time or under different 
circumstances, hence they are contextual (Helland 2018).

\section{Behavioral probability}

In real life, humans make decisions taking into account rational arguments, at the 
same time being influenced by irrational feelings and emotions. This implies that both 
quantities, the rational fraction and attraction factor define the probability $p(A_n)$ 
of choosing alternatives by decision makers. Thus the probability $p(A_n)$ of choosing 
an alternative $A_n$ embodies both a rational evaluation of the alternative utility 
as well as reflects the emotional attitude of decision makers towards the considered 
alternatives. This rational-irrational duality is typical for the behavior of real-life 
decision makers, because of which the probability $p(A_n)$ can be called 
{\it behavioral probability}. 

When looking for the form of this probability, it is necessary to keep in mind that 
the rational decision making has to be a particular case of the more general process 
encompassing both rational and irrational sides of decision making. That is, when 
irrational effects become not important, the choice becomes purely rational. This 
requirement can be written as a limiting condition.

\vskip 2mm

{\bf Correspondence principle}. Rational decision making is a particular case of 
behavioral decision making, when irrational effects play no role: 
\be
\label{17}
p(A_n) \ra f(A_n) \; ,  \qquad q(A_n) \ra 0 \;  .
\ee

\vskip 2mm

The behavior of decision makers reflects the superposition of rational and irrational 
sides of consciousness. In other words, the real-life behavior is a superposition of
cognition and emotions. This suggests the following axiom.

\vskip 2mm

{\bf Axiom 1}. Behavioral probability is the sum of a rational fraction and of an 
attraction factor:
\be
\label{18}   
p(A_n) = f(A_n) + q(A_n) \;   ,    
\ee
with $p(A_n)$ being semi-positive and normalized,
\be
\label{19}
\sum_{n=1}^{N_A} p(A_n) = 1 \; , \qquad 0 \leq p(A_n) \leq 1 \;  .
\ee
The condition of additivity is not required, so that, in general, the probability 
measure $\{p(A_n)\}$ is not necessarily additive.

\vskip 2mm

The rational part of the behavioral probability is explicitly defined by the rational 
fraction, while the irrational part is characterized by a quantity represented by 
the attraction factor. The behavioral probability, being a superposition of 
two terms, reflects the existence in life of rational-irrational duality, or 
cognition-emotion duality, or utility-attractiveness duality. When dealing with 
empirical data, the probability $p(A_n)$ describes the total fraction of decision 
makers preferring the given alternative. 

The alternatives $A_n$ from the set $\mathcal{A}$ acquire the following properties 
understood as the corresponding relations between their probabilities. 
  
\vskip 2mm 

\begin{enumerate}[label=(\roman*)]

\item
{\it Ordering}: For any two alternatives $A_1$ and $A_2$, one of the relations 
necessarily holds: either $A_1 \prec A_2$, in the sense that $p(A_1)<p(A_2)$, or 
$A_1\preceq A_2$, when $p(A_1) \leq p(A_2)$, or $A_1 \succ A_2$, if $p(A_1)>p(A_2)$,
or $A_1 \succeq A_2$, when $p(A_1)\geq p(A_2)$, or $A_1\sim A_2$, if $p(A_1)=p(A_2)$.

\item
{\it Linearity}: The relation $A_1 \preceq A_2$, implying $p(A_1) \leq p(A_2)$, means 
that $A_2 \succeq A_1$, in the sense that $p(A_2) \geq p(A_1)$. 

\item
{\it Transitivity}: For any three alternatives, such that $A_1 \preceq A_2$, with 
$p(A_1) \leq p(A_2)$, and $A_2 \preceq A_3$, when $p(A_2) \leq p(A_3)$, it follows 
that $A_1 \preceq A_3$, in the sense that $p(A_1) \leq p(A_3)$.

\item
{\it Completeness}: The set of alternatives $\cA$ contains a minimal $A_{min}$ and 
a maximal $A_{max}$ elements, for which $p(A_{min}) = \min_n p(A_n)$ and, respectively, 
$p(A_{max})=\max_n p(A_n)$. The ordered set of these alternatives is called a complete 
lattice. 

\end{enumerate}

\vskip 2mm

Relations between behavioral probabilities determine preference relations between 
the alternatives.

\vskip 2mm

{\bf Definition 6}. An alternative $A_1$ is called preferable to $A_2$ if and only 
if 
\be
\label{20}
 p(A_1) > p(A_2) \qquad  (A_1 \succ A_2) \; .
\ee  
Two alternatives $A_1$ and $A_2$ are indifferent if and only if
\be
\label{21}
 p(A_1) = p(A_2) \qquad  (A_1 \sim A_2) \; .
\ee

\vskip 2mm

{\bf Definition 7}. The alternative $A_{opt}$ is called optimal if and only if it 
corresponds to the maximal behavioral probability,
\be
\label{22}
p(A_{opt}) = \max_n p(A_n) \; .
\ee

\vskip 2mm

It is clear that an alternative can be more useful but not preferable, since its 
behavioral probability consists of a rational fraction and an irrational attraction 
factor. An alternative $A_1$ is preferable to $A_2$, implying that $p(A_1)>p(A_2)$, 
then and only then when 
\be
\label{23}
f(A_1) - f(A_2) > q(A_2) - q(A_1) \; .   
\ee
Both quantities, the rational fraction and attraction factor are important in the process 
of taking decisions.

\section{Attraction factor}

Although the attraction factor is a random quantity, it possesses, on average, some 
general properties that, because of their importance, are formulated as theorems.

\vskip 2mm

{\bf Theorem 1}. The attraction factor $q(A_n)$ varies in the interval
\be
\label{24}
- f(A_n) \leq q(A_n) \leq 1 - f(A_n) 
\ee 
and satisfies the {\it alternation law}
\be
\label{25}
\sum_{n=1}^{N_A} q(A_n) = 0 \;  .   
\ee 

\vskip 2mm

{\it Proof}. These properties follow directly from the definition of the behavioral
probability (\ref{18}), its semi-definiteness and normalization (\ref{19}), and from 
the semi-definiteness and normalization of the rational fraction (\ref{2}). $\square$

\vskip 2mm

Irrational feelings and emotions, playing a very important role in decision making,
are characterized by the attraction factor. Strictly speaking, the attraction factor 
$q(A_n)$ is a random quantity that varies for different people and different conditions. 
Despite that it is random, it enjoys some specific features that can be used for 
estimating the non-informative priors quantifying this factor.  

Recall that being random does not prevent the quantity from possessing well defined 
properties on average. In decision making this means that, although the attraction 
factor is difficult to define for a single decision maker and a single choice, but it 
may enjoy quite explicit properties at the aggregate level as an average for a large 
group of decision makers and over several choices. Such averages, playing the role of 
non-informative priors, could allow us to evaluate typical attraction factors, even 
having no detailed information on each of the separate decision makers.
 
\vskip 2mm

{\bf Definition 8}. If a quantity $y$ is given on an interval $[a,b]$, the average of 
$y$ is defined as the arithmetic average
\be
\label{26}
\overline y \equiv \frac{a+b}{2} \;   .
\ee
In the case when the boundaries $a$ and $b$ themselves are the quantities given on 
the intervals $[a_1, a_2]$ and, respectively, $[b_1, b_2]$, the non-informative prior 
for $y$ is the arithmetic average
\be
\label{27}
 \overline y \equiv \frac{\overline a + \overline b}{2} \qquad 
\left( \overline a \equiv \frac{a_1 + a_2}{2} \; , ~ 
       \overline b \equiv \frac{b_1 + b_2}{2} \right) \;  . 
\ee 

\vskip 2mm

{\bf Theorem 2}. {\it Quarter Law}. The average value for the attraction factor $q(A_n)$ 
in the positive quality class is
\be
\label{28}
 \overline q(A_n) = \frac{1}{4} \qquad ( A_n \in \cA_+ ) 
\ee
and in the negative quality class, it is 
\be
\label{29}
\overline q(A_n) = -\; \frac{1}{4} \qquad ( A_n \in \cA_- ) \;  .
\ee

\vskip 2mm

{\it Proof}. According to Theorem 1, the attraction factor is defined on the interval
$[-f(A_n), 1 - f(A_n)]$. Hence for the positive quality class it is given on the 
interval $[0, 1 - f(A_n)]$ and for the negative quality class, on the interval 
$[-f(A_n), 0]$. By the definition of the averages, we have for the positive quality class 
\be
\label{30}
\overline q(A_n) = \frac{1-\overline f(A_n)}{2} \qquad 
( A_n \in \cA_+ ) \; ,
\ee
while for the negative quality class
\be
\label{31}
 \overline q(A_n) = -\; \frac{\overline f(A_n)}{2} \qquad 
( A_n \in \cA_- ) \;  .
\ee
Since $f(A_n)$ is defined on the interval $[0,1]$, its non-informative prior is 
the average $\bar{f}(A_n) = 1/2$. This gives the averages (\ref{28}) and (\ref{29}). 
$\square$ 
 
\vskip 2mm

In this way, the average behavioral probability of choosing an alternative $A_n$ by 
a group of decision makers can be estimated by the expression
\be
\label{32}
p(A_n) = f(A_n) \pm 0.25 \;  ,
\ee
provided the probability properties (\ref{18}) are preserved.  

\vskip 2mm

The attraction factor, being a random quantity, varies for different agents as well 
as for the same decision maker at different moments of time. Therefore the same decision 
task, with the same utility factors, even for the same pool of subjects may be 
accompanied by different behavioral probabilities (Murphy and Fu 2018), hence different 
attraction factors. Such variations is a kind of random noise. Numerous empirical data 
show that these variations lead to statistical errors of about $0.1$ 
(Murphy and ten Brincke 2018). This implies that if the difference $p(A_n) - f(A_n)$ 
is smaller than $0.1$, then the attractiveness of the alternative pertains to the neutral 
class and the attraction factor can be set to zero.

\section{Multiple alternatives}

The estimate for the non-informative prior of the attraction factor, derived above,
is especially useful for the case of choosing between two alternatives. In the case, 
where there are many alternatives in the set $\mathcal{A}$, it is possible to more
precisely estimate typical attraction factors, playing the role of non-informative 
priors. 
   
Suppose $N_A$ alternatives can be classified according to the level of their 
attractiveness, so that 
\be 
\label{42}
 q(A_n) > q(A_{n+1}) \qquad ( n = 1,2,\ldots, N_A-1) \;  .
\ee
Let the nearest to each other attraction factors $q(A_n)$ and $q(A_{n+1})$ be separated 
by a typical gap 
\be
\label{43}  
\Dlt \equiv  q(A_n) - q(A_{n+1}) \; .
\ee
And let us accept that the average over the set $\mathcal{A}$ absolute value of the 
attraction factor can be estimated by the non-informative prior $\overline{q} = 1/4$, 
so that
\be
\label{44}
 \overline q \equiv \frac{1}{N_A} \sum_{n=1}^{N_A} |\; q(A_n)\; | = \frac{1}{4} \; .
\ee

\vskip 2mm

{\bf Theorem 3}. For a set $\mathcal{A}$ of $N_A$ alternatives, under conditions 
(\ref{42}), (\ref{43}), and (\ref{44}), the non-informative priors for the attraction 
factors are
$$
q(A_n) = \frac{N_A-2n+1}{2N_A} \qquad (N_A ~ even) \;  ,
$$
\be
\label{45}
 q(A_n) = \frac{N_A(N_A-2n+1)}{2(N_A^2-1)} \qquad (N_A ~ odd) \;  ,
\ee
depending on whether $N_A$ is even or odd.

\vskip 2mm

{\it Proof}. In accordance with conditions (\ref{42}), (\ref{43}), and (\ref{44}), we
can write 
\be
\label{46}
q(A_n) = q(A_1) - (n-1)\Dlt \; .
\ee
From the alternation law (\ref{25}), it follows
\be
\label{47}
 q(A_1) = \frac{N_A-1}{2} \; \Dlt \;  .
\ee
Using the definition of the average $\overline{q}$ in equation (\ref{44}) gives the gap
\begin{eqnarray}
\label{48}
\Dlt = \left\{ \begin{array}{ll}
4\overline q/N_A , ~ & N_A ~ even \\
4\overline q N_A/(N_A^2-1) , ~ & N_A ~ odd 
\end{array} \right.  \; ,
\end{eqnarray}
depending on whether the number of alternatives $N_A$ is even or odd. Then expression 
(\ref{47}) becomes
\begin{eqnarray}
\label{49}
q(A_1) = \left\{ \begin{array}{ll}
2\overline q(N_A-1)/N_A , ~ & N_A ~ even \\
2\overline q N_A/(N_A+1) , ~ & N_A ~ odd 
\end{array} \right.  \; .
\end{eqnarray}
Using (\ref{46}), we get
\begin{eqnarray}
\label{50}
q(A_n) = \left\{ \begin{array}{ll}
2\overline q(N_A+1-2n)/N_A , ~ & N_A ~ even \\
2\overline q N_A(N_A+1-2n)/(N_A^2-1) , ~ & N_A ~ odd 
\end{array} \right.  \; ,
\end{eqnarray}
In view of equality (\ref{44}), we have $\bar{q} = 1/4$. Then expression (\ref{49}) 
leads to
\begin{eqnarray}
\label{51}
q(A_1) = \left\{ \begin{array}{ll}
(N_A-1)/2N_A , ~ & N_A ~ even \\
N_A/2(N_A+1) , ~ & N_A ~ odd 
\end{array} \right.  \; .
\end{eqnarray}
And finally, equation (\ref{50}) results in the answer (\ref{45}). $\square$

\vskip 2mm

As applications of the above theorem, let us give some examples. Thus for a set of two 
alternatives, we have the already known values of the non-informative priors for the 
attraction factors
\be
\label{52}  
 \{ q(A_n) : ~ n = 1,2 \} = \left\{ \frac{1}{4}\;, \; - \; \frac{1}{4} \right\}\; .
\ee
For three alternatives, we find
\be
\label{53}
\{ q(A_n) : ~ n = 1,2,3 \} = \left\{ \frac{3}{8}\; , \; 0 \; , \; - \; \frac{3}{8} 
\right\}\; .
\ee
Respectively, for the set of four alternatives, we obtain the quality factors
\be
\label{54}
 \{ q(A_n) : ~ n = 1,2,3,4 \} = 
\left\{ \frac{3}{8}\; , \; \frac{1}{8}\; , \; - \; \frac{1}{8} \; , \; - \; \frac{3}{8} 
\right\} \; .
\ee

\section{Binary alternatives}

The case of binary alternatives is, probably, the most often considered in applications,
being a typical choice problem. In previous sections, a method of evaluating the average 
value of the attraction factors is described. As is shown, the typical attraction factor 
can be estimated, despite that, in general, attractiveness seems to be a vague notion. 
The natural question arises whether it would be feasible to give a more detailed 
assessment of attractiveness. This problem has been discussed for the case of two 
alternatives, when it has been necessary to choose between two lotteries 
(Favre et al. 2016; Vincent et al. 2016; Ferro et al. 2021, Zhang and Kjellstr\"{o}m 2021).

Let us consider the choice between two lotteries, $A$ and $B$, whose utilities are 
$U(A)$ and $U(B)$, respectively. These quantities can represent either the standard 
expected utility (von Neumann and Morgenstern 1953) or other utility functionals employed 
in decision theory, e.g. the utility functional of prospect theory 
(Kahneman and Tversky 1979; Tversky and Kahneman 1992). The attraction factor in the form
\be
\label{a1}
q(A) = \min\{ \vp(A), \; \vp(B)\} \tanh\{ a [\; U(A) - U(B)\;] \} \; , 
\qquad q(B) = - q(A) \; ,
\ee
has been considered (Vincent et al. 2016; Ferro et al. 2021), where
$$
 \vp(A) = \frac{1}{Z} \; e^{\bt U(A)} \; , \qquad 
 \vp(B) = \frac{1}{Z} \; e^{\bt U(B)} \; , \qquad 
Z = e^{\bt U(A)} + e^{\bt U(B)} \; .
$$
It has been used for characterizing the series of $91$ binary decision tasks (lotteries), 
where the choice is made by the pool of $142$ subjects (Murphy and ten Brincke 2018). 
The parameters of the attraction factor are fitted so that to optimally agree with the 
given experimental data. It is shown (Vincent et al. 2016; Ferro et al. 2021) that the 
decision theory with this attraction factor better describes the empirical data than the 
stochastic cumulative prospect theory (Tversky and Kahneman 1992) and than the stochastic 
rank-dependent utility theory (Quiggin 1982).

As is clear, the parameters calibrated so that to optimally describe a given set of 
lotteries may be not appropriate for another set of lotteries. In that sense, each 
combination of parameters is contextual, being suitable for a particular set of decision 
tasks, but not necessarily adequate for other groups of decision problems. This can be 
easily understood noticing that the attraction factor (\ref{a1}) becomes zero, when the 
utilities $U(A)$ and $U(B)$ coincide, or it becomes negligible when these utilities are 
close to each other. At the same time very close, or equal utilities often correspond 
to high uncertainty in the choice, which results in large attraction factors. A typical 
example of such a situation has been illustrated by Kahneman and Tversky (1979) for a 
set of lotteries with close or coinciding expected utilities.     

The choice between two alternatives with equal or close utilities is a kind of the 
"Buridan's donkey problem" (Kane 2005) that can be solved only considering emotions.

\section{Buridan's donkey problem}

Emotions are characterized by attraction factors. Hence to quantify emotions means 
the necessity of evaluating the values of the attraction factors for the considered 
alternatives. As is explained in Sec. 5, the first step in this estimation is the 
formulation of the quarter law stating that the non-informative prior for the magnitude 
of the attraction factor for either positive or negative quality classes is $\pm 0.25$. 
However we need to define how the actual classification could be realized, so that 
each alternative would be associated with the corresponding quality class, either 
positive or negative. This problem is typical for the research area known as soft 
computing aspiring to find methods that tolerate imprecision and uncertainty of fuzzy 
notions to achieve tractability and robustness allowing for quantitative conclusions 
(Clocksin 2003; de Silva 2003; Jamshidi 2003). 
 
Below we suggest an algorithm that is applicable to that situation of close utilities
of lotteries, whether with gains or with losses. This algorithm can be justified on 
the basis of studies in experimental neuroscience, which have discovered that, when 
making a choice, the main and foremost attention of decision makers is directed towards 
the payoff probabilities (Kim, Seligman and Kable 2012). This implies that subjects 
evaluate higher the probabilities than the related payoffs (Yukalov and Sornette 2014, 2018). 
In mathematical terms, this can be formulated as the existence of different types of 
scaling for the alternative quality with respect to payoff utility and payoff 
probability. Say, the payoff utility is scaled linearly, while the payoff probability, 
exponentially. 

To be explicit, let us consider a simple case of two alternatives, one lottery
\be
\label{33}
A_1 = \{u,p \; | \; 0, 1 - p \} \;  ,
\ee
with a payoff utility $u$ and a related probability $p$, and the other lottery
\be
\label{34}
A_2 = \left\{\lbd u, \frac{p}{\lbd} \; | \; 0, 1 - \;\frac{p}{\lbd} \right\} \; ,
\ee
whose payoff utility and probability are scaled in such a way that the expected 
utilities of both lotteries are equal,
$$
 U(A_1) = U(A_2) = up  \; .
$$
Then the corresponding rational fractions coincide, $f(A_1) = f(A_2) = 1/2$, and 
one cannot choose a preferable alternative being based on rational arguments. This 
is a typical example of a series of lotteries considered by Kahneman and Tversky 
(1979). However, empirical studies show that subjects do make clear preferences 
between the lotteries, depending on their payoffs and probabilities. This implies 
that subjects are able to intuitively classify the lotteries into positive 
(attractive) or negative (repulsive). 

Since "quality" or "attractiveness" are vague notions, it would be tempting to 
accomplish the quality classification by means of words. For instance, we could accept 
that between two lotteries that one is of better quality, or more attractive, that 
yields a more certain gain or less certain loss. Often this is a reasonable way of 
classification, although not always.

Suppose the lottery $A_1$ is quite certain, which implies that the payoff probability 
is in the interval $1/2 < p \leq 1$. Hence the average probability is $p = 3/4$, which 
is appreciated by people as highly certain (Hillson 2003, 2019). Let the scaling with 
$\lbd > 1$ be such that the payoff utility increases, while its probability 
diminishes. When $\lbd$ is not large, subjects do prefer the more certain lottery 
$A_1$. However strongly increasing the payoff utility may attract more people, 
despite a small payoff probability, as has been confirmed by real-life lotteries 
(Rabin 2000). 

In order to describe the method of classification of alternatives into positive 
(attractive) or negative (repulsive) quality classes, let us introduce the 
{\it quality functional} $Q(A_n)$. The fact that decision makers in their choice pay 
the main and foremost attention to the payoff probabilities (Kim, Seligman and 
Kable 2012) is formalized by a linear dependence of the quality functional with 
respect to the payoff utility and by an exponential dependence with respect to the 
payoff probability. For the case of the lotteries (\ref{33}) and (\ref{34}), this 
implies the quality functionals
$$
Q(A_1) = u b^p \; , \qquad Q(A_2) = \lbd u b^{p/\lbd} \;  .
$$
When the scaling $\lambda$ is of order one and $A_1$ is more certain, subjects 
consider the more certain lottery as more attractive, which means that $Q(A_1)$ is 
larger than $Q(A_2)$. But if the payoff probability is diminished by an order, which
assumes $\lbd=10$, while the payoff utility increases by an order, then the lottery 
$A_1$ can become less attractive than $A_2$, so that $Q(A_1)$ becomes smaller than 
$Q(A_2)$. The change of attractiveness occurs where $Q(A_1)=Q(A_2)$. The latter 
equality gives the expression for the base $b$ that for $p=3/4$ and $\lbd=10$ yields
\be
\label{35}
 b = \lbd^{\lbd/(\lbd-1)p} =  30 \;  .
\ee
The above arguments give a clue allowing us to define the quality functional for any 
lottery.

\vskip 2mm

{\bf Definition 9}. The quality functional of an arbitrary lottery is
\be
\label{36}
 Q(A_n) = \sum_i u(x_i) 30^{p_n(x_i)} \;  .
\ee

\vskip 2mm

Comparing the quality functionals of different lotteries, we can meet the case, where
these functionals are equal, but the lotteries differ from each other by the gain-loss
number difference
\be
\label{37}
N(A_n) = N_+(A_n) - N_-(A_n)
\ee 
between the number of admissible gains $N_+(A_n)$ and the number of possible losses 
$N_-(A_n)$. A typical example is the comparison of the lottery $A_1$ defined in 
(\ref{33}) and the lottery
\be
\label{38}
A_3 = \{ u_1,p \; | \; u_2, p \; | \; 0 , 1 - 2p \} \;  ,
\ee
in which $u_1 + u_2 = u$. Then the related quality functionals are equal
$$
 Q(A_3) = u_1 b^p + u_2 b^p = u b^p = Q(A_1)  \; ,
$$
where $b=30$. If $u_n>0$, then the lottery $A_1$ possesses only one admissible gain
and no losses, while the lottery $A_3$, two gains and also no losses. Hence $N(A_1)=1$ 
and $N(A_3)=2$. Since $N(A_3)$ is larger than $N(A_1)$, the lottery $A_3$ is treated 
as more attractive. Similarly, for the lotteries with losses, where $u_n<0$, and 
quality functionals are equal, the gain-loss number difference $N(A_1)=-1$ is larger 
than $N(A_3)=-2$, so that the lottery $A_1$ with a smaller number of losses is more 
attractive.  

\vskip 2mm 

{\bf Definition 10}. A lottery $A_1$ is of better quality, or more attractive, than 
$A_2$, so that $q(A_1) > q(A_2)$, if either
\be
\label{39}
 Q(A_1) > Q(A_2) \;  ,   
\ee
or if
\be
\label{40}
  Q(A_1) = Q(A_2) \; , \qquad  N(A_1) > N(A_2) \; .
\ee

If some lotteries $A_1$ and $A_2$ cannot be classified as more or less attractive, they 
are said to be of equal quality, or equally attractive, so that $q(A_1)=q(A_2)$. If there
are only two of these lotteries, then the alternation law $q(A_1) + q(A_2) = 0$ implies 
$q(A_1) = q(A_2) = 0$. In that case, the lotteries are in the neutral quality class.  

\vskip 2mm

If the alternative $A_1$ is more attractive than $A_2$, then the related behavioral 
probabilities can be estimated as 
\be
\label{41}
p(A_1) = f(A_1) + 0.25 \; , \qquad p(A_2) = f(A_2) - 0.25 \;  ,
\ee
where $f(A_n)$ are rational fractions. Here the inequality $0\leq p(A_n)\leq 1$ is 
assumed, which can be formalized by the definition
$$
p(A_n) = {\rm Ret}_{[0,1]} \{ f(A_n) \pm  0.25 \} \; ,
$$
where the retract function is defined as
\begin{eqnarray}
\nonumber
{\rm Ret}_{[0,1]}z = \left\{  \begin{array}{ll}
0 , ~ & ~ z < 0 \\
z , ~ & ~ 0 \leq z \leq 1 \\
1 , ~ & ~ z > 1 
\end{array} \right. .
\end{eqnarray}

Recall that the above expressions estimate the aggregate fractions of decision makers 
averaged over many subjects and a set of choices. For a single decision maker, the 
attraction factor is a random quantity. However, the average attraction factor and, 
respectively, the average behavioral probability can be estimated according to rules 
(\ref{41}). The rational fraction $f(A_n)$ shows the fraction (frequentist probability) 
of decision makers that would choose the corresponding alternative on the basis of 
only rational rules. While the behavioral probability $p(A_n)$ defines the real total 
fraction of decision makers actually choosing $A_n$, taking into account both the 
rational utility as well as the irrational emotional attractiveness of the alternatives.

\section{Kahneman-Tversky lotteries}

By a number of examples, Kahneman and Tversky (1979) have shown that the expected 
utility theory in many cases does not work at all, so that decision makers do not 
decide according to utility theory, because of the very close lottery utilities, and 
even choose the alternatives that should be neglected according to the utility theory 
prescriptions. In their experiments, the number of participants was about $100$. The 
typical statistical error was close to $\pm 0.1$. Payoffs below are given in monetary 
units, whose measures are of no importance when using dimensionless rational fractions.   

Below, we show that the method described above correctly predicts the aggregate choice, 
giving good quantitative estimates for behavioral probabilities. Rational fractions 
are calculated by formulas of Sec. 2. For simplicity, the linear utility function $u(x)$ 
is accepted. The attraction factor is represented by its non-informative prior, with the 
sign prescribed by the lottery quality functional defined in Sec. 8. For brevity, we 
use the notation $Q(A_n) \equiv Q_n$. 

For the convenience of the reader, we summarize the formulae that are used below in 
characterizing the lotteries. The rational utility fraction is calculated according to
the definition in Sec. 2 as 
$$
f(L_n) = \frac{U(L_n)}{\sum_n U(L_n)} \qquad 
( U(L_n) \geq 0 )
$$ 
for semi-positive expected utilities and as
$$
f(L_n) = \frac{|\;U(L_n)\;|^{-1}}{\sum_n|\; U(L_n)\;|^{-1}} \qquad 
( U(L_n) < 0 )
$$
for negative expected utilities, where the latter are given by the expression
$$    
U(L_n) = \sum_i x_i p_n(x_i) \; .
$$
The quality functional $Q_n=Q(L_n)$ is defined in (\ref{36}). 

\vskip 2mm

{\it Choice 1}. Consider two lotteries
$$
L_1 = \{ 2.5, \; 0.33 \; | \; 2.4, \; 0.66 \; | \; 0, \; 0.01 \} \; , \qquad
L_2 = \{2.4 , \; 1 \} \; .
$$
The rational fractions are $f(L_1) = 0.501$ and $f(L_2) = 0.499$, so that the first 
lottery should be chosen on the rational grounds. However, the lottery quality 
functionals $Q_1 = 30.3$ and $Q_2 = 72$ show that the second lottery, being more 
certain, is more attractive, since $Q_2 > Q_1$. Hence $q(L_2) > q(L_1)$, and involving 
the non-informative prior, we have $q(L_1) = - 0.25$, while $q(L_2) = 0.25$. This 
gives the behavioral probabilities
$$
p(L_1) = 0.25 \; , \qquad p(L_2) = 0.75 \; ,
$$
according to which the second lottery is optimal. This is in agreement with the 
empirical results
$$
p_{exp}(L_1) = 0.18 \; , \qquad p_{exp}(L_2) = 0.82 \; .
$$
The more certain, but less useful lottery is chosen. 

\vskip 2mm

{\it Choice 2}. One chooses between the lotteries
$$
L_1 = \{ 2.5, \; 0.33 \; | \; 0, \; 0.67 \} \; , \qquad 
L_2 = \{ 2.4, \; 0.34 \; | \; 0, \; 0.66 \} \;   .
$$
The rational fractions are close to each other, $f(L_1) = 0.503$ and $f(L_2)=0.497$.
At the first glance, it is difficult to say which of the lotteries is more attractive, 
since the first lottery has a slightly larger payoff, while the second is a little 
more certain. But the lottery qualities $Q_1 = 7.68$ and $Q_2 = 7.63$ show that the 
first lottery is a bit more attractive. Hence $q(L_1)=0.25$ and $q(L_2)=-0.25$. Then 
the behavioral probabilities are
$$
p(L_1) = 0.75 \; , \qquad p(L_2) = 0.25
$$
which is well comparable with the experimental data
$$
 p_{exp}(L_1) = 0.83 \; , \qquad p_{exp}(L_2) = 0.17 \;  .
$$
This is an example, where the majority prefer a less certain, but more useful 
lottery.  

\vskip 2mm

{\it Choice 3}. Considering the lotteries
$$
L_1 = \{ 4, \; 0.8 \; | \; 0, \; 0.2 \} \; , \qquad 
L_2 = \{ 3, \; 1 \} \;  ,
$$
one sees that the first lottery, although being less certain, is more useful, having 
a larger rational fraction $f(L_1) = 0.516$ while $f(L_2) = 0.484$. But its quality 
is lower than that of the second lottery, $Q_1 = 60.8$, while $Q_2 = 90$. This means 
that the second lottery is more attractive, because of which $q(L_1)=-0.25$ and 
$q(L_2)=0.25$. As a result, the behavioral probabilities are
$$
 p(L_1) = 0.27 \; , \qquad p(L_2) = 0.73 \;  ,
$$
being close to the experimentally observed
$$
 p_{exp}(L_1) = 0.20 \; , \qquad p_{exp}(L_2) = 0.80 \;   .
$$
Here the more certain, although less useful lottery is chosen.  

\vskip 2mm

{\it Choice 4}. For the lotteries
$$
L_1 = \{ 4, \; 0.20 \; | \; 0, \; 0.80 \} \; , \qquad 
L_2 = \{ 3, \; 0.25 \; | \; 0 , \; 0.75 \} \;  ,
$$
the rational fractions are again close to each other, as in the previous case, 
$f(L_1) = 0.516$ and $f(L_2) = 0.484$. But the quality of the first lottery is higher 
than that of the second, $Q_1 = 7.9$, but $Q_2 = 7.02$. This makes the first lottery 
more attractive, with $q(L_1) = 0.25$ and $q(L_2) = - 0.25$. And the choice reverses, 
as compared to the previous case,
$$
 p(L_1) = 0.77 \; , \qquad p(L_2) = 0.23 \;   ,
$$  
in agreement with the empirical results
$$
p_{exp}(L_1) = 0.65 \; , \qquad p_{exp}(L_2) = 0.35 \;   .
$$
Again, a less certain, although more useful, lottery is chosen.

\vskip 5mm

{\it Choice 5}. Between the lotteries
$$
L_1 = \{ 6, \; 0.45 \; | \; 0, \; 0.55 \} \; , \qquad 
L_2 = \{ 3, \; 0.9 \; | \; 0 , \; 0.10 \} \;     ,
$$
it is difficult to choose which is better. The first lottery suggests a twice larger 
payoff, while the second, twice higher payoff probability. The utility of both the 
lotteries is the same, with the same rational fractions $f(L_1)= f(L_2)=0.5$. However, 
the lottery qualities are different, $Q_1=27.7$, while $Q_2=64.1$, showing that the 
second lottery is more attractive, which gives $q(L_1) = - 0.25$ and $q(L_2) = 0.25$. 
Therefore the behavioral probabilities become
$$     
p(L_1) = 0.25 \; , \qquad p(L_2) = 0.75 \; .
$$
And the empirical data are
$$
p_{exp}(L_1) = 0.14 \; , \qquad p_{exp}(L_2) = 0.86 \; .
$$
More certain lottery is chosen. 

\vskip 2mm

{\it Choice 6}. The lotteries
$$
L_1 = \{ 6, \; 0.001 \; | \; 0, \; 0.999 \} \; , \qquad 
L_2 = \{ 3, \; 0.002 \; | \; 0 , \; 0.998 \} \; ,
$$
have the same rational fractions $f(L_1)=f(L_2)=0.5$. But the quality of the first 
lottery is higher than that of the second, $Q_1 = 6.02$, while $Q_2 = 3.02$. That is, 
the first lottery is more attractive, so that $q(L_1) = 0.25$ and $q(L_2) = - 0.25$.
This yields the behavioral probabilities
$$
p(L_1) = 0.75 \; , \qquad p(L_2) = 0.25 \;   ,
$$
practically coinciding with the experimental data
$$
p_{exp}(L_1) = 0.73 \; , \qquad p_{exp}(L_2) = 0.27 \; .
$$
Between two equally useful lotteries, the less certain is chosen.

\vskip 2mm

{\it Choice 7}. For the lotteries
$$
L_1 = \{ 6, \; 0.25 \; | \; 0, \; 0.75 \} \; , \qquad 
L_2 = \{ 4, \; 0.25 \; | \; 2 , \; 0.25 \; | \; 0, \; 0.5 \} \;   ,
$$
the rational fractions are again the same, which does not make it possible to 
choose on the basis of utility, $f(L_1)=f(L_2)=0.5$. Although the lottery qualities 
are equal, $Q_1=Q_2=14$, but the second lottery suggests a larger choice of gains, 
$N(L_2) = 2 > N(L_1) = 1$, which makes it more attractive, with $q(L_1) = - 0.25$ and 
$q(L_2) = 0.25$. As a result, the behavioral probabilities read as
$$
p(L_1) = 0.25 \; , \qquad p(L_2) = 0.75 \; .
$$
The empirical data are  
$$
p_{exp}(L_1) = 0.18 \; , \qquad p_{exp}(L_2) = 0.82 \; .
$$
Among seemingly equally useful lotteries, the choice is made under the influence of 
the attraction factor.  

\vskip 2mm

{\it Choice 8}. Considering the lotteries
$$
L_1 = \{ 5, \; 0.001 \; | \; 0, \; 0.999 \} \; , \qquad 
L_2 = \{ 0.005, \; 1 \} \; ,
$$
we see that they are of equal utility, with the rational fractions $f(L_1)=f(L_2)=0.5$.
But the lottery qualities are essentially different, $Q_1=5.02$ and $Q_2=0.15$, which 
defines the attraction factors $q(L_1) = 0.25$ and $q(L_2) = - 0.25$. Then the behavioral 
probabilities are
$$
p(L_1) = 0.75 \; , \qquad p(L_2) = 0.25 \;  .
$$
This is very close to the empirical data
$$
p_{exp}(L_1) = 0.72 \; , \qquad p_{exp}(L_2) = 0.28 \; .
$$
Again, this is an example, when a less certain lottery is chosen among two equally 
useful lotteries. 

\vskip 2mm

{\it Choice 9}. The lotteries
$$
L_1 = \{ 10, \; 0.5 \; | \; 0, \; 0.5 \} \; , \qquad 
L_2 = \{ 5, \; 1 \} 
$$
have equal utilities, with the rational fractions $f(L_1)=f(L_2)=0.5$. The lottery
qualities $Q_1=54.8$ and $Q_2=150$ show that the second lottery is more attractive,
hence $q(L_1)=-0.25$ and $q(L_2)=0.25$. Therefore the behavioral probabilities become
$$
p(L_1) = 0.25 \; , \qquad p(L_2) = 0.75 \; .
$$
The empirical probabilities are
$$
p_{exp}(L_1) = 0.16 \; , \qquad p_{exp}(L_2) = 0.84 \; .
$$
The second lottery is more certain, although has a smaller payoff.

\vskip 2mm

{\it Choice 10}. For the lotteries
$$
L_1 = \{ 2, \; 0.5 \; | \; 1, \; 0.5 \} \; , \qquad 
L_2 = \{ 1.5, \; 1 \} \; ,
$$
rational fractions are equal, $f(L_1) = f(L_2) = 0.5$. The lottery qualities are
$Q_1 = 16.4$ and $Q_2 = 45$. Hence the second lottery is more attractive, which means
that $q(L_1) = - 0.25$ and $q(L_2) = 0.25$. Then the behavioral probabilities are
$$
p(L_1) = 0.25 \; , \qquad p(L_2) = 0.75 \; .
$$
This is very close to the experimentally found probabilities
$$
p_{exp}(L_1) = 0.20 \; , \qquad p_{exp}(L_2) = 0.80 \; ,
$$
actually coinciding with them within the accuracy of experiments. 

\vskip 2mm

{\it Choice 11}. The previous lotteries dealt with gains. Now we shall treat the 
lotteries with losses, which implies that the subject has to pay, that is to loose,
the amount of monetary units marked as negative. Consider the lotteries
$$
L_1 = \{ -4, \; 0.8 \; | \; 0, \; 0.2 \} \; , \qquad 
L_2 = \{ -3, \; 1 \} \;   .
$$ 
The rational fraction of the second lottery is larger, $f(L_1) = 0.484$, while 
$f(L_2) = 0.516$. However, the first lottery is more attractive, since its quality is 
higher, $Q_1 = - 60.8$, while $Q_2 = - 90$. This tells us that $q(L_1) = 0.25$ and 
$q(L_2) = - 0.25$, which leads to the behavioral probabilities
$$
p(L_1) = 0.73 \; , \qquad p(L_2) = 0.27 \;  .
$$
In experiments, the majority also choose the first lottery,
$$
p_{exp}(L_1) = 0.92 \; , \qquad p_{exp}(L_2) = 0.08 \;   .
$$
The situation is opposite to the case of gains. Now a lottery with a less certain 
loss is preferable. 

\vskip 2mm

{\it Choice 12}. For the lotteries
$$
L_1 = \{ -4, \; 0.2 \; | \; 0, \; 0.8 \} \; , \qquad 
L_2 = \{ -3, \; 0.25 \; | \; 0 , \; 0.75  \} \;  ,
$$
the rational fractions are $f(L_1)=0.484$ and $f(L_2)=0.516$. The related lottery 
qualities read as $Q_1 =-7.9$ and $Q_2 =-7.02$, showing that the second lottery is 
more attractive, with $q(L_1)=-0.25$ and $q(L_2)=0.25$. Then we find the behavioral 
probabilities
$$
p(L_1) = 0.23 \; , \qquad p(L_2) = 0.77 \;   .
$$
Now the majority of decision makers choose the second lottery,
$$
p_{exp}(L_1) = 0.42 \; , \qquad p_{exp}(L_2) = 0.58 \; ,
$$
although it suggests a more certain loss. 
 
\vskip 2mm

{\it Choice 13}. The lotteries
$$
L_1 = \{ -3, \; 0.9 \; | \; 0, \; 0.1 \} \; , \qquad 
L_2 = \{ -6, \; 0.45 \; | \; 0 , \; 0.55  \} 
$$
possess equal utility, hence equal rational fractions $f(L_1) =  f(L_2) = 0.5$.
But the second lottery is more attractive, since its quality is higher,
$Q_1 = - 64.1$, while $Q_2 = - 27.7$. Therefore $q(L_1)=-0.25$ and $q(L_2)=0.25$. 
The behavioral probabilities
$$
p(L_1) = 0.25 \; , \qquad p(L_2) = 0.75 
$$
show that the second lottery is optimal, in agreement with the empirical observations,
$$
p_{exp}(L_1) = 0.08 \; , \qquad p_{exp}(L_2) = 0.92 \; .
$$
The second lottery is preferred, although its loss is higher, but the loss is less 
certain. 

\vskip 2mm

{\it Choice 14}. For the lotteries
$$
L_1 = \{ -3, \; 0.002 \; | \; 0, \; 0.998 \} \; , \qquad 
L_2 = \{ -6, \; 0.001 \; | \; 0 , \; 0.999  \}   
$$
rational fractions are equal, $f(L_1)=f(L_2)=0.5$. However, the lottery qualities
$Q_1=-3.02$ and $Q_2=-6.02$ demonstrate that the first lottery is more attractive, 
so that $q(L_1) = 0.25$ and $q(L_2) = - 0.25$. This results in the behavioral 
probabilities
$$
p(L_1) = 0.75 \; , \qquad p(L_2) = 0.25  
$$
that are very close to the experimentally found,
$$
p_{exp}(L_1) = 0.70 \; , \qquad p_{exp}(L_2) = 0.30 \;   .
$$
Now, between two equally useful lotteries, the lottery suggesting a more certain 
loss is chosen.

\vskip 2mm

{\it Choice 15}. The lotteries
$$
L_1 = \{ -1, \; 0.5 \; | \; 0, \; 0.5 \} \; , \qquad 
L_2 = \{ -0.5, \; 1  \} 
$$
also have equal rational fractions, $f(L_1) =  f(L_2) = 0.5$. But the quality of 
the first lottery is higher, $Q_1 = - 5.48$, while $Q_2 = - 15$. Hence the first 
lottery is more attractive, with $q(L_1) = 0.25$, but $q(L_2) = - 0.25$. The 
resulting behavioral probabilities 
$$
 p(L_1) = 0.75 \; , \qquad p(L_2) = 0.25 
$$
are in good agreement with the experimental data
$$
p_{exp}(L_1) = 0.69 \; , \qquad p_{exp}(L_2) = 0.31 \;   .
$$
The first lottery is preferred, although its loss is larger. 

\vskip 2mm

{\it Choice 16}. Among the lotteries
$$
L_1 = \{ -6, \; 0.25 \; | \; 0, \; 0.75 \} \; , \qquad 
L_2 = \{ -4, \; 0.25 \; | \; -2 , \; 0.25 \; | \; 0, \; 0.5 \} \;   ,
$$
that look similar, having the same rational fractions $f(L_1) =  f(L_2) = 0.5$, 
and equal qualities $Q_1 =  Q_2 = - 14$, the second is less attractive, exhibiting 
a larger number of losses, $N(L_1) = - 1 > N(L_2) = -2$. Therefore $q(L_1)=0.25$ 
and $q(L_2)=-0.25$. This yields the behavioral probabilities
$$
 p(L_1) = 0.75 \; , \qquad p(L_2) = 0.25 \;  ,
$$ 
practically coinciding with the empirical data
$$
p_{exp}(L_1) = 0.70 \; , \qquad p_{exp}(L_2) = 0.30 \; ,
$$
within the accuracy of experiments.

\vskip 2mm

{\it Choice 17}. The lotteries
$$
L_1 = \{ -5, \; 0.001 \; | \; 0, \; 0.999 \} \; , \qquad 
L_2 = \{ -0.005, \; 1 \} 
$$
have the same utility, with equal rational fractions $f(L_1) =  f(L_2) = 0.5$.
But their qualities $Q_1 = - 5.02$ and $Q_2 = - 0.15$ show that the second lottery 
is more attractive, having a much larger quality. Hence $q(L_1) = - 0.25$ and 
$q(L_2) = 0.25$. The behavioral probabilities are
$$
 p(L_1) = 0.25 \; , \qquad p(L_2) = 0.75 \;  ,
$$
as compared with the experimental data
$$
p_{exp}(L_1) = 0.17 \; , \qquad p_{exp}(L_2) = 0.83 \;   .
$$
Surprisingly, the lottery with certain loss is chosen, which is explained by its 
higher quality. 

\vskip 2mm

{\it Choice 18}. For the lotteries
$$
L_1 = \{ -10, \; 0.5 \; | \; 0, \; 0.5 \} \; , \qquad 
L_2 = \{ -5, \; 1  \}  \; ,
$$
the rational fractions are equal, $f(L_1)=f(L_2)=0.5$. But for the lottery qualities,
we have $Q_1=-54.8$ and $Q_2=-150$. Thus the first lottery is more attractive, hence
$q(L_1) = 0.25$ and $q(L_2) = - 0.25$. This gives the behavioral probabilities
$$
 p(L_1) = 0.75 \; , \qquad p(L_2) = 0.25 \;  ,
$$ 
in agreement with empirical data
$$
p_{exp}(L_1) = 0.69 \; , \qquad p_{exp}(L_2) = 0.31 \; .
$$
Now the lottery with a larger, but less certain loss is chosen. 

\vskip 2mm

The results for all $18$ choices between the Kahneman-Tversky lotteries are summarized 
in Table 1 showing which of the lotteries is optimal, that is, having the largest 
predicted behavioral probability 
\be
\label{55}
 p(L_{opt}) \equiv \max_n p(L_n)  
\ee
over the given lattice of alternatives. Also shown are the rational fractions for 
the optimal lottery, $f(L_{opt})$, experimental probabilities of the optimal lottery, 
defined as the fractions of decision makers choosing the optimal lottery 
$p_{exp}(L_{opt})$, and the related empirical attraction factors 
\be
\label{56}
q_{exp}(L_{opt}) = p_{exp}(L_{opt}) - f(L_{opt}) \; .
\ee
The results, corresponding to the non-optimal lotteries, can be easily found from the 
normalization conditions
\be
\label{57}
p(L_1) + p(L_2) = 1 \; , \qquad f(L_1) + f(L_2) = 1 \; , \qquad  
q(L_1) + q(L_2) = 0 \;  .
\ee
At the bottom of Table 1, the average values over all $18$ cases are given for the 
rational fraction $\overline f(L_{opt}) = 0.5$, predicted behavioral probability
$\overline p(L_{opt}) = 0.75$, experimentally observed probability 
$\overline p_{exp}(L_{opt}) = 0.77$, and the experimentally observed average absolute 
value of the attraction factor
\be
\label{58} 
\overline q_{exp} = \overline p_{exp}(L_{opt}) - \overline f(L_{opt}) = 0.27 \;  .
\ee
Within the accuracy of the experiment, the predicted average behavioral probability 
of choosing an optimal lottery, $0.75$, equals the empirical average fraction of 
decision makers $0.77$, and the average attraction factor $0.27$ practically coincides 
with the theoretical estimate of $0.25$.

As the analysis of this set of choices demonstrates, it is not possible to predict 
the behavioral decision making of humans by considering separately either lottery 
utilities, payoffs, or payoff probabilities. But reliable predictions can be made by 
defining behavioral probabilities, including the estimates of both, rational fractions 
as well as attraction factors. On the aggregate level, such predictions are not merely 
qualitative, but provide good quantitative agreement with empirical data, involving 
no fitting parameters.      

At the same time, the expected utility theory is not applicable to the Kahneman-Tversky 
lotteries, since the lottery with a higher utility is preferred only twice among
$18$ lotteries, that is only in the $1/9$ part of the lotteries. Also, it is important 
to notice that the formula (\ref{a1}) here is not valid, as far as for the coinciding 
utilities it gives zero attraction factor, while the aggregate experimental data give 
for the attraction factor $0.27$.

\section{Quarter law}

In the previous sections, it has been shown that the average influence of emotions 
in decision making can be quantified by the typical value of attraction factor,
which turns out to be close to $0.25$, which is termed {\it quarter law} and which
follows from the non-informative prior estimate of Sec. 5. Thus in the set of 
Kahneman-Tversky lotteries of Sec. 9 the experimentally measured average attraction 
factor is $0.27$, which, within the typical statistical error of $0.1$, coincides with
the predicted attraction factor $0.25$. 

In the present section, we verify the quarter law on the basis of a large set of binary 
lotteries studied recently (Murphy and ten Brincke 2018). In the analyzed experiment, 
$142$ subjects were suggested a set of binary decision tasks (lotteries). The same 
experiment was repeated after two weeks, with randomly changing the order of the pairs 
of lotteries. The experiments at these two different times are referred as session $1$
and session $2$. There are three types of lotteries: lotteries containing only gains
(all payoffs are positive), lotteries with only losses (all payoffs are negative), and 
mixed lotteries containing gains as well as losses. As usual, a loss implies the 
necessity to pay the designed amount of money. Keeping in mind the estimation of 
attraction factors in positive and negative quality classes, we consider the related 
lotteries, where the difference between the rational utility factors and the empirical 
choice probabilities, at least in one of the sessions, are larger than the value of 
the typical statistical error of $0.1$ corresponding to random noise. On the basis of 
these lotteries, we calculate the quantities of interest and summarize the results in 
several tables.

Table 2 presents the results for the optimal lotteries with only gains and Table 3 
shows the results for the optimal lotteries with only losses. Recall that a lottery 
$L_1$ is called optimal, as compared to a lottery $L_2$ if and only if the corresponding 
probability $p(L_1)$ is larger than $p(L_2)$. In both the cases, of either the lotteries 
with only gains or the lotteries with only losses, an optimal lottery is always a lottery
from the positive quality class, in which $q(L_{opt})>0$. The situation can be different
for the mixed lotteries, containing gains as well as losses. In these cases, an optimal 
lottery can occasionally pertain to a negative quality class. Table 4 summarizes the 
results for the mixed lotteries containing both gains and losses. Among these lotteries,
the first sixteen examples in Table 4 are the lotteries from the positive quality class, 
which at the same time are the optimal lotteries. The last five cases are the lotteries
that are not optimal, however being from the positive quality class. 

As is seen, the value of the attraction factor in the positive or negative quality 
classes is $\pm 0.22$, which is in very good agreement with the predicted non-informative
priors $\pm 0.25$. Thus the quarter law provides a rather accurate estimate of the 
attraction factor at the aggregate level.

\section{Conclusion}

An approach is developed allowing for the quantification of emotions in decision 
making. The approach takes into account the duality of decision making, including 
both rational and irrational sides of decision process. The rational evaluation 
of alternatives is based on logical clearly prescribed rules defining a rational 
fraction representing the probability of choosing alternatives on the basis of 
rational principles. 

The irrational side of decision processes is due to subconscious feelings, emotions, 
and intuition that cannot be exactly measured for a given subject at a given moment 
of time, thus inducing emotional uncertainty in the process of decision making. 
Irrational processes are superimposed on the rational evaluation of the considered 
alternatives and define for each alternative a correction term called attraction factor. 
Since irrational processes cannot be exactly quantified, the attraction factor is a 
random quantity. The attraction factor can be described by linguistic characteristics 
that can be classified into three quality classes, briefly speaking, positive, negative, 
and neutral. The positive quality class includes such specifications as attractive, 
pleasant, good and like that. The negative quality class comprises the features like 
repulsive, unpleasant, bad, and so on. The neutral quality class is intermediate, 
being neither positive nor negative. 

The attraction factor is a variable randomly varying for different decision makers 
and even for the same decision maker at different moments of time. Nevertheless, being 
random, does not preclude this quantity to have a well defined average value inside 
each of the quality classes. A theorem is proved defining the average values of the 
quality factor inside the positive class as $1/4$ and inside the negative class 
as $-1/4$. For alternatives represented by lotteries with equal or close utilities, 
a method is suggested ascribing each lottery to the appropriate quality class. 

Being able to determine the belonging of alternatives to the related quality 
classes and knowing the average values of attraction factors allows us to find the 
average behavioral probabilities associated with the typical fractions of decision 
makers choosing this or that alternative. 

The method is illustrated by a series of lotteries with a difficult choice, when 
the standard expected utility theory is not applicable, or its prescriptions 
contradict the choice of real humans. The empirical data confirm that the 
non-informative prior for attraction factors provides an accurate quantification
of emotions at the aggregate level.

Summarizing, the main points of the suggested approach can be formulated as follows. 

\vskip 2mm
\begin{enumerate}[label=(\roman*)]
\item
Decision making is treated as a probabilistic process that can be characterized by
behavioral probabilities defining the portions of decision makers choosing this or 
that alternative from the given set of alternatives. 

\item
The behavioral probability, taking into account the rational-irrational or
cognition-emotion duality of decision processes, describes decision making affected
by emotions. The superposition of utility and attractiveness is represented as a sum 
of two terms, a rational fraction and an attraction factor. 

\item
The rational fraction, having the properties of the standard additive probability, 
describes the fraction of decision makers that would make their choice being based 
solely on rational grounds, following prescribed rational rules. The rational 
fraction quantifies the utility of the choice.

\item
The attraction factor takes into account irrational effects influencing the choice, 
such as feelings, emotions, and biases. The attraction factor characterizes 
subconscious attractiveness of the considered alternatives, because of which it is 
called attraction factor. The attraction factor is a random quantity, varying for 
different subjects, different choices, and different times.  

\item
Despite being random, the attraction factor possesses well defined average features. 
The average values of the attraction factor for positive or negative quality classes
can be defined by non-informative priors. 

\item
The approach makes it possible to give quantitative predictions in the choice 
between the lotteries with emotional uncertainty, where the expected utility theory 
does not work. The aggregate predictions, averaged over decision makers and choices, 
are in good quantitative agreement with empirical data. 

\item
Empirical data confirm the quarter law providing, at the aggregate level, an accurate 
evaluation of typical influence of emotions in decision making.

\item
The appealing feature of the approach is its straightforward axiomatic formulation 
employing rather simple mathematics. Although the structure of the approach is 
implicitly influenced by quantum theory, but it completely avoids borrowed from physics 
complicated quantum techniques.

\end{enumerate}

\section*{Acknowledgment}

The author is grateful for helpful advise and useful discussions to D. Sornette and 
E.P. Yukalova. 

\vskip 2mm

This research did not receive any specific grant from funding agencies in the public,
commercial, or not-for-profit sector. 

\vskip 15mm

\section*{References}

{\parindent=0pt

\vskip 2mm
N.I. Al-Najjar, J. Weinstein (2009)
The ambiguity aversion literature: a critical assessment, 
Econ. Philos. 25: 249--284.

\vskip 2mm
N.I. Al-Najjar, J. Weinstein (2009)  
The ambiguity aversion literature: a critical assessment, 
Econ. Philos. 25: 357--369.

\vskip 2mm
M. Amjadzadeh, K. Ansari-Asl (2017)
An innovative emotion assessment using physiological signals based on the combination 
mechanism,
Scientia Iranica D 24: 3157--3170.

\vskip 2mm
D. Ariely (2008)
Predictably Irrational, 
Harper, New York.

\vskip 2mm
M.H. Birnbaum (2008) 
New paradoxes of risky decision making,
Psychol. Rev. 115: 463--501. 

\vskip 2mm
W.F. Clocksin (2003)
Artificial intelligence and the future,
Phil. Trans. Roy. Soc. Lond. A  361: 1721--1748. 

\vskip 2mm
C.W. de Silva (2003) 
The role of soft computing in intelligent machines, 
Phil. Trans. Roy. Soc. Lond. A 361: 1749--1780.

\vskip 2mm
M. Favre, A. Wittwer, H.R. Heinimann, V.I. Yukalov, D. Sornette (2016)
Quantum decision theory in simple risky choices,
PLOS One 11: 0168045. 

\vskip 2mm
G.M. Ferro, T. Kovalenko, D. Sornette (2021)
Quantum decision theory augments rank-dependent expected utility and cumulative prospect 
theory,
J. Econ. Psychol. 86: 102417.

\vskip 2mm
M. Jamshidi (2003) 
Tools for intelligent control: fuzzy controllers, neural networks and genetic 
algorithms, 
Phil. Trans. Roy. Soc. Lond. A 361: 1781--1808.

\vskip 2mm
C. Julmi (2019) 
When rational decision-making becomes irrational: a critical assessment and 
re-conceptualization of intuition effectiveness,
Business Research 12: 291--314.

\vskip 2mm
I.S. Helland (2018) 
Epistemic Processes, 
Springer, Cham.

\vskip 2mm
D. Hillson (2003) 
Effective Opportunity Management for Projects,
Marcel Dekker, New York.

\vskip 2mm
D. Hillson (2019) 
Capturing Upside Risk,
CRC Press, Boca Raton.

\vskip 2mm
D. Kahneman (1982)
Judgment under Uncertainty, Heuristics and Biases,
Cambridge University, Cambridge.

\vskip 2mm
D. Kahneman, A. Tversky (1979) 
Prospect theory: an analysis of decision under risk.
Econometrica 47: 263--292.

\vskip 2mm
R. Kane (2005) 
A Contemporary Introduction to Free Will, 
Oxford University, New York.

\vskip 2mm
B.E. Kim, D. Seligman, J.M. Kable (2012) 
Preference reversals in decision making under risk are accompanied by changes in 
attention to different attributes, 
Front. Neurosci. 6: 109. 

\vskip 2mm
R.D. Luce (1959)  
Individual Choice Behavior: A Theoretical Analysis, 
Wiley, New York. 

\vskip 2mm
R.D. Luce, R. Raiffa (1989) 
Games and Decisions: Introduction and Critical Survey, 
Dover, New York.

\vskip 2mm
M.J. Machina (2008) 
Non-expected utility theory, in: New Palgrave Dictionary of Economics, 
eds. S.N. Durlauf, L.E. Blume, Macmillan, New York.

\vskip 2mm
A.D. Milner, M.A. Goodale (2008)
Two visual systems re-viewed,
Neuropsychologia 46: 774--785.

\vskip 2mm
M. Minsky (2006)
The Emotion Machine, 
Simon and Schuster, New York.

\vskip 2mm
A. Murphy, L. Fu (2018)
The effect of confidence in valuation estimates on arbitrager behavior and market prices,
J. Behav. Finance 19: 349--363.

\vskip 2mm
R.O. Murphy, R.H.W. ten Brincke (2018)
Hierarchical maximum likelihood parameter estimation for cumulative prospect theory:
Improving the reliability of individual risk parameter estimates,
Management Sci. 64: 308--326.

\vskip 2mm
R.E. Neapolitan, X. Jiang (2018)
Artificial Intelligence,
CRC Press, Boca Raton. 

\vskip 2mm
R. Picard (1997)
Affective Computing,
Massachusetts Institute of Technology, Cambridge.

\vskip 2mm
H. Plessner, C. Betsch, T. Betsch (2008)
Intuition in Judgment and Decision Making, 
Lawrence Erlbaum Associates, New York. 

\vskip 2mm
D.L. Poole, A.K. Mackworth (2017)
Artificial Intelligence,
Cambridge University, Cambridge. 

\vskip 2mm
J. Quiggin (1982)
A theory of anticipated utility,
J. Econ. Behav. Org. 3: 323--343.

\vskip 2mm
M. Rabin (2000)
Risk aversion and expected-utility theory: a calibration theorem,
Econometrica 68: 1281--1292.

\vskip 2mm
S.J. Russel, P. Norvig (2016)
Artificial Intelligence: A Modern Approach,
Pearson Education, Harlow.

\vskip 2mm
Z. Safra, U. Segal (2008) 
Calibration results for non-expected utility theories,
Econometrica 76: 1143--1166.

\vskip 2mm
L.J. Savage (1954)
The Foundations of Statistics,
Wiley, New York.

\vskip 2mm
K.R. Scherer, A. Moors (2019)
The emotion process: Event appraisal and component differentiation, 
Ann. Rev. Psychol. 70: 719--745. 

\vskip 2mm
J.R. Searle (2001)  
Rationality in Action, 
Massachusetts Institute of Technology, Cambridge.

\vskip 2mm
P. Slovic, A. Tversky (1974)
Who accepts Savage's axioms?
Behav. Sci. 19: 368--373. 

\vskip 2mm
A. Tversky, D. Kahneman (1992)
Advances in prospect theory: Cumulative representation of uncertainty,
J. Risk Uncert. 5: 297--323. 

\vskip 2mm
A.V. Vartanov, I.I. Vartanova (2018)
Four-dimensional spherical model of emotion,
Procedia Computer Science 145: 604--610.

\vskip 2mm
A. Vartanov, V. Ivanov, I. Vartanova (2020)
Facial expressions and subjective assessments of emotions,
Cogn. Syst. Res. 59: 319--328.

\vskip 2mm
S. Vincent, T. Kovalenko, V.I. Yukalov, D. Sornette (2016)
Calibration of quantum decision theory: Aversion to large losses and predictability 
of probabilistic choices, \\
http://ssrn.com/abstract=2775279.

\vskip 2mm
J. von Neumann, O. Morgenstern (1953) 
Theory of Games and Economic Behavior, 
Princeton University, Princeton.

\vskip 2mm
L. Wang, H.Y. Liu, W.L. Liang, T.H. Zhou (2021)
Emotional expression analysis based on fine-grade emotion quantification model
via social media,  
in Advances in Intelligent Information Hiding and Multimedia Signal Processing, 
eds. J.S. Pan et al., p. 211--218, Springer, Singapore.  

\vskip 2mm
V.I. Yukalov, D. Sornette (2008)
Quantum decision theory as quantum theory of measurement, 
Phys. Lett. A 372: 6867--6871.

\vskip 2mm
V.I. Yukalov, D. Sornette (2009)
Scheme of thinking quantum systems, 
Laser Phys. Lett. 6: 833--839.

\vskip 2mm
V.I. Yukalov, D. Sornette (2009)
Physics of risk and uncertainty in quantum decision making,
Eur. Phys. J. B 71: 533--548.

\vskip 2mm
V.I. Yukalov, D. Sornette (2011) 
Decision theory with prospect interference and entanglement,
Theory Decis. 70: 283--328.

\vskip 2mm
V.I. Yukalov, D. Sornette (2014) 
Manipulating decision making of typical agents, 
IEEE Trans. Syst. Man Cybern. Syst. 44: 1155--1168.

\vskip 2mm
V.I. Yukalov, D. Sornette (2016)
Quantum probability and quantum decision making, 
Philos. Trans. Roy. Soc. A 374: 20150100.

\vskip 2mm
V.I. Yukalov, D. Sornette (2018) 
Quantitative predictions in quantum decision theory, 
IEEE Trans. Syst. Man Cybern. Syst. 48: 366--381.  

\vskip 2mm
V.I. Yukalov (2020)
Evolutionary processes in quantum decision theory,
Entropy 22: 681.

\vskip 2mm
V.I. Yukalov (2021)
Tossing quantum coins and dice,
Laser Phys. 31: 055201.

\vskip 2mm
V.I. Yukalov (2021)
A resolution of St. Petersburg paradox,
J. Math. Econ. \\ https://doi.org/10.1016/j.jmateco.2021.102537.

\vskip 2mm
M. Zafirovski (2012)
Beneath rational choice: Elements of irrational choice theory,
Current Sociol. 61: 3--21.

\vskip 2mm
C. Zhang, H. Kjellstr\"{o}m (2021)
A subjective model of human decision making based on quantum decision theory,
arXiv: 2101.05851. 

}

\newpage

\begin{table}[hp]
%Table 1
\caption{Optimal lotteries $L_{opt}$ from the Kahneman-Tversky set, the rational 
fractions for the optimal lotteries, $f(L_{opt})$, predicted behavioral probabilities 
$p(L_{opt})$, experimentally observed probabilities $p_{exp}(L_{opt})$, defined as the 
fractions of the participants choosing the optimal lottery $L_{opt}$, and the experimental 
attraction factors $q_{exp}(L_{opt})$ corresponding to the optimal lotteries. At the 
bottom, the average values are shown.}

\vskip 5mm

\centering
\renewcommand{\arraystretch}{1.2}
\begin{tabular}{|c|c|c|c|c|c|} \hline
   & $L_{opt}$ & $f(L_{opt})$  & $p(L_{opt})$ & $p_{exp}(L_{opt})$ &  $q_{exp}(L_{opt})$ \\ \hline
1  &   $L_2$   & 0.50  & 0.75 & 0.82   & 0.32 \\ 
2  &   $L_1$   & 0.50  & 0.75 & 0.83   & 0.33 \\ 
3  &   $L_2$   & 0.48  & 0.73 & 0.80   & 0.32 \\ 
4  &   $L_1$   & 0.52  & 0.77 & 0.65   & 0.13 \\
5  &   $L_2$   & 0.50  & 0.75 & 0.86   & 0.36  \\ 
6  &   $L_1$   & 0.50  & 0.75 & 0.73   & 0.23 \\
7  &   $L_2$   & 0.50  & 0.75 & 0.82   & 0.32 \\
8  &   $L_1$   & 0.50  & 0.75 & 0.72   & 0.22 \\ 
9  &   $L_2$   & 0.50  & 0.75 & 0.84   & 0.34 \\ 
10 &   $L_2$   & 0.50  & 0.75 & 0.80   & 0.30 \\ 
11 &   $L_1$   & 0.48  & 0.73 & 0.92   & 0.44 \\
12 &   $L_2$   & 0.52  & 0.77 & 0.58   & 0.06 \\
13 &   $L_2$   & 0.50  & 0.75 & 0.92   & 0.42 \\
14 &   $L_1$   & 0.50  & 0.75 & 0.70   & 0.20 \\ 
15 &   $L_1$   & 0.50  & 0.75 & 0.69   & 0.19 \\
16 &   $L_1$   & 0.50  & 0.75 & 0.70   & 0.20 \\
17 &   $L_2$   & 0.50  & 0.75 & 0.83   & 0.33 \\
18 &   $L_1$   & 0.50  & 0.75 & 0.69   & 0.19 \\  \hline
   &           & 0.50  & 0.75 & 0.77   & 0.27 \\ \hline
\end{tabular}
\end{table}

\begin{table}[hp]
%Table 2
\caption{Optimal lotteries with gains. The rational fraction $f(L_{opt})$ of the 
optimal lottery, fractions of subjects (frequentist probabilities) $p_i(L_{opt})$ 
choosing the optimal lottery in the session $i=1,2$, and the attraction factors 
$q_i(L_{opt})$ of the optimal lottery in the session $i$. At the bottom, the average 
values for the related quantities.  }

\vskip 5mm
\centering
\renewcommand{\arraystretch}{1.2}
\begin{tabular}{|c|c|c|c|c|c|} \hline
   & $f(L_{opt})$ & $p_1(L_{opt})$  & $p_2(L_{opt})$ & $q_1(L_{opt})$ & $q_2(L_{opt})$ \\ \hline
1  &   0.55       & 0.86 & 0.89 & 0.31   & 0.34 \\ 
2  &   0.48       & 0.66 & 0.69 & 0.18   & 0.21 \\ 
3  &   0.51       & 0.68 & 0.62 & 0.17   & 0.11 \\ 
4  &   0.59       & 0.80 & 0.75 & 0.22   & 0.17 \\
5  &   0.63       & 0.89 & 0.90 & 0.26   & 0.27  \\ 
6  &   0.66       & 0.96 & 0.95 & 0.30   & 0.29 \\
7  &   0.51       & 0.79 & 0.81 & 0.28   & 0.30 \\
8  &   0.48       & 0.60 & 0.63 & 0.12   & 0.15 \\ 
9  &   0.63       & 0.88 & 0.92 & 0.26   & 0.30 \\ 
10 &   0.56       & 0.89 & 0.82 & 0.33   & 0.26 \\ 
11 &   0.63       & 0.77 & 0.73 & 0.14   & 0.10 \\
12 &   0.51       & 0.72 & 0.73 & 0.21   & 0.21 \\
13 &   0.61       & 0.87 & 0.85 & 0.26   & 0.24 \\
14 &   0.63       & 0.93 & 0.93 & 0.30   & 0.30 \\ 
15 &   0.64       & 0.85 & 0.87 & 0.21   & 0.23 \\
16 &   0.64       & 0.80 & 0.80 & 0.16   & 0.16 \\
17 &   0.64       & 0.89 & 0.89 & 0.25   & 0.25 \\
18 &   0.48       & 0.65 & 0.70 & 0.17   & 0.22 \\  
19 &   0.65       & 0.87 & 0.93 & 0.22   & 0.28 \\
20 &   0.66       & 0.86 & 0.82 & 0.20   & 0.16 \\
21 &   0.58       & 0.84 & 0.80 & 0.26   & 0.22 \\
22 &   0.52       & 0.75 & 0.74 & 0.23   & 0.22 \\
23 &   0.48       & 0.64 & 0.65 & 0.16   & 0.17 \\
24 &   0.44       & 0.60 & 0.53 & 0.16   & 0.10 \\
25 &   0.62       & 0.73 & 0.79 & 0.11   & 0.17 \\
26 &   0.64       & 0.81 & 0.90 & 0.17   & 0.26 \\
27 &   0.66       & 0.93 & 0.96 & 0.27   & 0.30 \\ \hline
   &   0.58       & 0.80 & 0.80 & 0.22   & 0.22 \\ \hline
\end{tabular}
\end{table}

\newpage

\begin{table}[hp]
%Table 3
\caption{Optimal lotteries with losses. The rational fraction $f(L_{opt})$ of the 
optimal lottery, fractions of subjects (frequentist probabilities) $p_i(L_{opt})$ 
choosing the optimal lottery in the session $i=1,2$, and the attraction factors 
$q_i(L_{opt})$ of the optimal lottery in the session $i$. At the bottom, the average 
values for the related quantities. }

\vskip 5mm
\centering
\renewcommand{\arraystretch}{1.2}
\begin{tabular}{|c|c|c|c|c|c|} \hline
   & $f(L_{opt})$ & $p_1(L_{opt})$  & $p_2(L_{opt})$ & $q_1(L_{opt})$ & $q_2(L_{opt})$ \\ \hline
1  &   0.52       & 0.77 & 0.75 & 0.25   & 0.23 \\ 
2  &   0.60       & 0.85 & 0.83 & 0.25   & 0.23 \\ 
3  &   0.53       & 0.72 & 0.71 & 0.19   & 0.18 \\ 
4  &   0.64       & 0.96 & 0.92 & 0.32   & 0.28 \\
5  &   0.55       & 0.70 & 0.68 & 0.15   & 0.13  \\ 
6  &   0.54       & 0.73 & 0.72 & 0.20   & 0.19 \\
7  &   0.63       & 0.79 & 0.84 & 0.16   & 0.21 \\
8  &   0.54       & 0.66 & 0.63 & 0.12   & 0.09 \\ 
9  &   0.56       & 0.80 & 0.89 & 0.24   & 0.33 \\ 
10 &   0.58       & 0.89 & 0.92 & 0.31   & 0.34 \\ 
11 &   0.49       & 0.66 & 0.71 & 0.17   & 0.22 \\
12 &   0.62       & 0.87 & 0.93 & 0.25   & 0.31 \\
13 &   0.55       & 0.79 & 0.74 & 0.24   & 0.19 \\
14 &   0.54       & 0.82 & 0.77 & 0.29   & 0.24 \\ 
15 &   0.53       & 0.65 & 0.70 & 0.12   & 0.17 \\
16 &   0.51       & 0.59 & 0.62 & 0.08   & 0.11 \\
17 &   0.56       & 0.79 & 0.86 & 0.23   & 0.30 \\
18 &   0.58       & 0.89 & 0.90 & 0.31   & 0.32 \\  
19 &   0.61       & 0.76 & 0.74 & 0.15   & 0.13 \\ \hline
   &   0.56       & 0.77 & 0.78 & 0.21   & 0.22 \\ \hline
\end{tabular}
\end{table}

\newpage

\begin{table}[hp]
%Table 4
\caption{Mixed lotteries, containing gains and losses, from the positive quality 
class. The rational fraction $f(L_+)$ of the lottery, fractions of subjects $p_i(L_+)$ 
choosing the corresponding lottery in the session $i=1,2$, and the attraction factors
$q_i(L_+)$ of the lottery in that session $i$. At the bottom, the average values of the
related quantities. }

\vskip 5mm
\centering
\renewcommand{\arraystretch}{1.2}
\begin{tabular}{|c|c|c|c|c|c|} \hline
   & $f(L_+)$     & $p_1(L_+)$  & $p_2(L_+)$ & $q_1(L_+)$ & $q_2(L_+)$ \\ \hline
1  &   0.40       & 0.69 & 0.66 & 0.29   & 0.26 \\ 
2  &   0.62       & 0.85 & 0.85 & 0.23   & 0.23 \\ 
3  &   0.67       & 0.87 & 0.82 & 0.20   & 0.15 \\ 
4  &   0.44       & 0.62 & 0.61 & 0.18   & 0.17 \\
5  &   0.50       & 0.64 & 0.54 & 0.15   & 0.05  \\ 
6  &   0.59       & 0.71 & 0.65 & 0.12   & 0.06 \\
7  &   0.54       & 0.69 & 0.63 & 0.16   & 0.10 \\
8  &   0.49       & 0.66 & 0.60 & 0.18   & 0.16 \\ 
9  &   0.57       & 0.87 & 0.85 & 0.30   & 0.28 \\ 
10 &   0.65       & 0.75 & 0.77 & 0.10   & 0.12 \\ 
11 &   0.52       & 0.77 & 0.70 & 0.26   & 0.19 \\
12 &   0.49       & 0.58 & 0.63 & 0.09   & 0.14 \\
13 &   0.55       & 0.87 & 0.92 & 0.32   & 0.37 \\
14 &   0.52       & 0.61 & 0.67 & 0.09   & 0.15 \\ 
15 &   0.53       & 0.80 & 0.83 & 0.27   & 0.30 \\
16 &   0.56       & 0.67 & 0.63 & 0.11   & 0.07 \\
17 &   0.00       & 0.27 & 0.27 & 0.27   & 0.27 \\
18 &   0.00       & 0.29 & 0.36 & 0.29   & 0.36 \\  
19 &   0.00       & 0.30 & 0.45 & 0.30   & 0.45 \\
20 &   0.00       & 0.39 & 0.38 & 0.39   & 0.38 \\
21 &   0.00       & 0.37 & 0.35 & 0.37   & 0.35 \\ \hline
   &   0.41       & 0.63 & 0.63 & 0.22   & 0.22 \\ \hline
\end{tabular}
\end{table}

\end{document}